\documentclass{article}

\usepackage{arxiv}
\usepackage{subcaption}
\usepackage[utf8]{inputenc} 
\usepackage[T1]{fontenc}    
\usepackage{hyperref}       
\usepackage{url}            
\usepackage{booktabs}       
\usepackage{amsfonts}       
\usepackage{nicefrac}       
\usepackage{microtype}      
\usepackage{lipsum}
\usepackage{amsmath,amsfonts}
\usepackage{xcolor}
\usepackage{pifont}  

\usepackage{graphicx}
\graphicspath{ {./images/} }
\usepackage[utf8]{inputenc} 
\usepackage[T1]{fontenc}    
\usepackage{hyperref}       
\usepackage{url}            
\usepackage{booktabs}       
\usepackage{amsfonts}       
\usepackage{nicefrac}       
\usepackage{microtype}      
\usepackage{lipsum}
\usepackage{graphicx}
\graphicspath{ {./images/} }
\usepackage{xcolor}
\usepackage{amsmath}
\usepackage{tcolorbox}
\definecolor{RubineRed}{RGB}{209, 0, 86} 
\definecolor{BurntOrange}{RGB}{204, 85, 0}
\definecolor{lightblue}{rgb}{0.68, 0.85, 0.9} 
\definecolor{steelblue}{RGB}{70, 130, 180} 
\definecolor{electricblue}{RGB}{5, 60, 131}
\definecolor{lightblue2}{RGB}{ 8, 105, 231}
\definecolor{lightorange}{RGB}{225, 144, 79 }
\definecolor{turqoise}{RGB}{79, 225, 201}
\definecolor{greenlight}{RGB}{3, 146, 24}
\definecolor{Orange}{RGB}{56, 39, 0}

\title{Hybrid Reasoning Based on Large Language Models for Autonomous Car Driving}

\author{
Mehdi Azarafza \\
Department of Computer Science \\
 Univ. of Applied Science Hamm-Lippstadt \\
  Carl von Ossietzky University of Oldenburg\\
  Lippstadt, Germany\\
  \texttt{mehdi.azarafza@hshl.de} \\
   \And
Mojtaba Nayyeri \\
Institute For Artificial Intelligence \\
 University of Stuttgart\\
  Stuttgart, Germany\\
  \texttt{mojtaba.nayyeri@ipvs.uni-stuttgart.de} \\
  \And
 Charles Steinmetz \\
Department of Computer Science \\
 Univ. of Applied Science Hamm-Lippstadt \\
  Carl von Ossietzky University of Oldenburg\\
  Lippstadt, Germany\\
  \texttt{charles.steinmetz@hshl.de} \\
  \And
 Steffen Staab \\
Institute For Artificial Intelligence \\
 University of Stuttgart\\
  Stuttgart, Germany\\
  \texttt{Steffen.Staab@ipvs.uni-stuttgart.de} \\
  \And
 Achim Rettberg \\
Department of Computer Science \\
 Univ. of Applied Science Hamm-Lippstadt\\
 Carl von Ossietzky University of Oldenburg\\
  Lippstadt, Germany\\
  \texttt{achim.rettberg@hshl.de}  \\
}

\begin{document}
\maketitle
\begin{abstract}
Large Language Models (LLMs) have garnered significant attention for their ability to understand text and images, generate human-like text, and perform complex reasoning tasks.
However, their ability to generalize this advanced reasoning with a combination of natural language text for decision-making in dynamic situations requires further exploration. 
In this study, we investigate how well LLMs can adapt and apply a combination of arithmetic and common-sense reasoning, particularly in autonomous driving scenarios. 
We hypothesize that LLMs' hybrid reasoning abilities can improve autonomous driving by enabling them to analyze detected object and sensor data, understand driving regulations and physical laws, and offer additional context.
This addresses complex scenarios, like decisions in low visibility (due to weather conditions), where traditional methods might fall short. 
We evaluated Large Language Models (LLMs) based on accuracy by comparing their answers with human-generated ground truth inside CARLA. The results showed that when a combination of images (detected objects) and sensor data is fed into the LLM, it can offer precise information for brake and throttle control in autonomous vehicles across various weather conditions. This formulation and answers can assist in decision-making for auto-pilot systems. 
\end{abstract}


\section{Introduction}
Over past decade, there has been a notable acceleration in the progress of autonomous driving vehicle technology, mainly driven by significant advancements in the field of deep learning and artificial intelligence (AI).
Autonomous driving requires the ability to make decisions in dynamic and unpredictable situations \cite{decision-making-car}.
The decision-making technique in autonomous driving can be guided either by optimization-based rules such as lane-keeping control or by data-driven machine intelligence like reinforcement learning \cite{survery-dm}. 
While rule-based approaches are often criticized for their limited \textit{generalizability} \ref{tab1} in complex traffic situations \cite{join-sadat, survery-dm}, data-driven approaches like deep learning and reinforcement learning have shown better performance in certain intricate scenarios \cite{CIRL,survery-dm}.  

\begin{figure}[htbp]
\includegraphics[width=.5\linewidth, height=7cm]{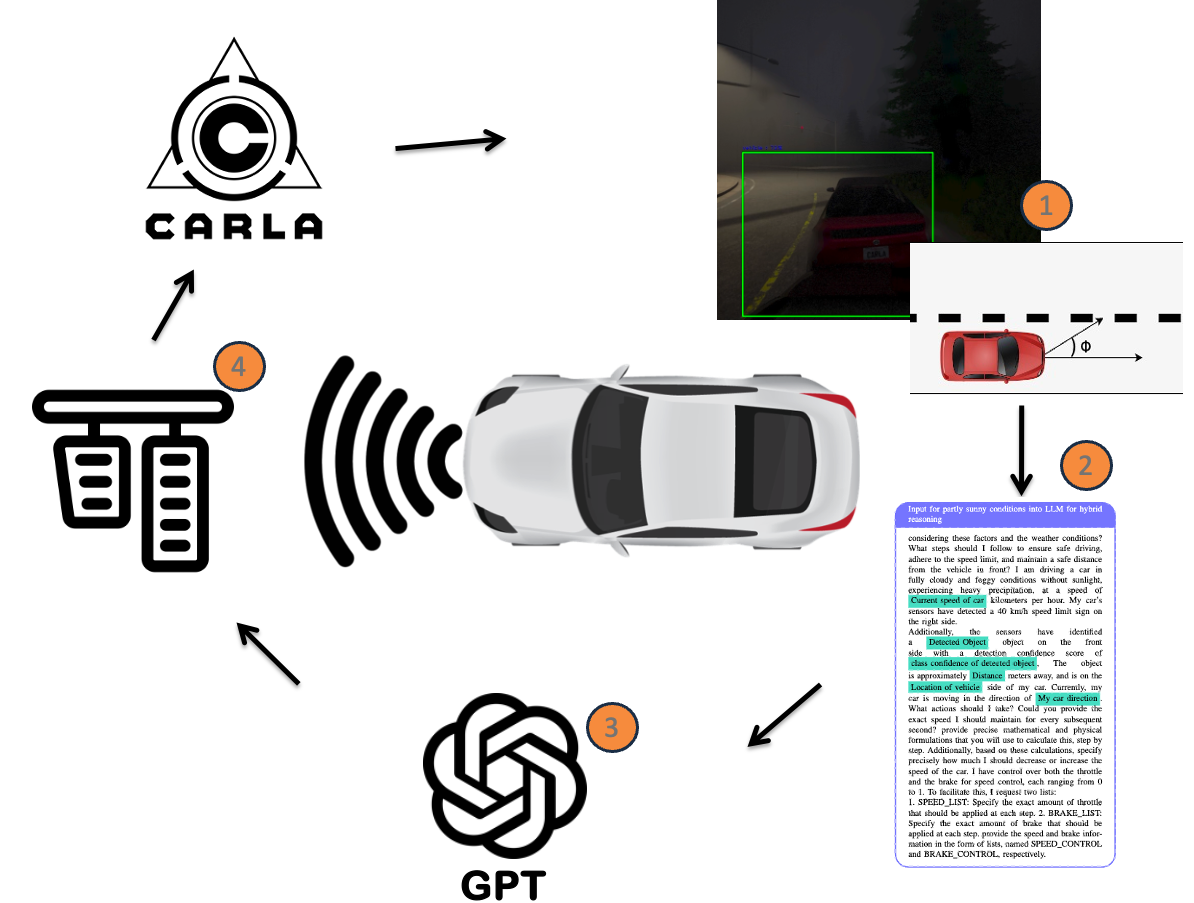} 
\centering
\caption{Workflow of hybrid reasoning of LLM in CARLA}
\label{fig:workflow}
\end{figure}

Despite the achievements of current autonomous driving methods, particularly reinforcement learning (RL) has some challenges.
One such challenge is the absence of formal guarantees regarding agent behavior (based on rewards\cite{survey-reasoning2024}), making it difficult to ensure consistently correct decisions (\textit{Decision-making} \ref{tab1}), particularly in complex scenarios that require sophisticated reasoning.
While rule-based systems rely on predefined traffic laws crafted by human experts, and RL is specifically tailored for certain tasks\cite{survey-reasoning2024}. LLMs can consider a multitude of factors, including real-time sensor inputs, detected objects, environmental conditions, direction of the car and location of car, and then make decisions based on all of these parameters. LLM can process and integrate various types of information simultaneously to make informed decisions (\textit{Comprehensive Awareness} \ref{tab1}). \\
As outlined in Table \ref{tab1}, each model has specific limitations: Rule-based methods, although capable of real-time adaptation, are restricted by their dependence on predefined rules and scenarios, which makes them less effective in highly dynamic environments. RL excels in such environments \cite{survey-reasoning2024} due to its capacity for learning and adaptation based on feedback; however, RL methods often struggle with generalizability\cite{survey-reasoning2024}. Large Language Models (LLMs) perform well in dynamic settings and are more easily generalizable, but they may encounter issues with response time, potentially limiting their effectiveness in real-time applications.

To address these limitations, we suggest developing a robust, domain-specific small language model that is trained on structured inputs to produce a single command at a time for the vehicle. While this paper focuses on the reasoning abilities of LLMs for autonomous driving, the task of optimizing language models for real-time performance is left for future work.
\begin{table}[htbp]
\caption{A General Comparison of Reasoning abilities with Large Language Models, Reinforcement
Learning and Rule Based}
\begin{center}
\begin{tabular}{|c|c|c|c|}
\hline
\textbf{Criteria}&\multicolumn{3}{|c|}{\textbf{Methods}} \\
\cline{2-4} 
\textbf{} & \textbf{\textit{LLM$^{\mathrm{a}}$}}& \textbf{\textit{RL$^{\mathrm{b}}$}}& \textbf{\textit{Rule-based}} \\
\hline
Decision-making &\textcolor{green}{\checkmark}&  \textcolor{green}{\checkmark}&  \textcolor{green}{\checkmark}\\
Flexibility &\textcolor{green}{\checkmark}& {\color{red}\ding{55}}&{\color{red}\ding{55}} \\
Comprehensive Awareness &\textcolor{green}{\checkmark}
& {\color{red}\ding{55}}
&{\color{red}\ding{55}}
\\
Generalization &\textcolor{green}{\checkmark}
&{\color{red}\ding{55}} 
&{\color{red}\ding{55}}
\\
Real-time Response &  {\color{red}\ding{55}}
& \textcolor{green}{\checkmark}
&\textcolor{green}{\checkmark}
\\

\hline
\multicolumn{4}{l}{$^{\mathrm{a}}$Large Language Model, $^{\mathrm{b}}$Reinforcement learning
}
\end{tabular}
\label{tab1}
\end{center}
\end{table}
Given these challenges, we investigate how Large Language Models (LLMs), particularly their intricate reasoning capabilities
can serve as a complementary assistance to improve decision-making in autonomous driving.

Large Language Models (LLMs) employ an optimized transformer architecture in an auto-regressive manner\cite{transformers}. These models are primarily trained on an extensive dataset containing trillions of tokens and incorporating billions of parameters. Subsequently, they undergo supervised fine-tuning and reinforcement learning with human feedback (RLHF) to align with human preferences, emphasizing both helpfulness and safety.
Recent investigations have showcased LLMs' capacity to tackle intricate tasks such as arithmetic, commonsense reasoning, and symbolic reasoning, is achieved through an approach termed Chain-of-thought prompting~\cite{chain-of-thought}. 

This paper aims to assess and investigate hybrid reasoning, specifically in the context of arithmetic within commonsense reasoning for autonomous vehicle decision-making with LLM inside CARLA\cite{carla} as run-time evaluation under various meteorological conditions.
To achieve this goal, we adopt a prompting approach anchored in object detection. 
Our decision-making reasoning workflow is depicted in Figure ~\ref{fig:workflow}. 
In hybrid reasoning, combining arithmetic and commonsense elements, we utilize the objects detected by YOLOv8, illustrated as part 1 in Figure \ref{fig:workflow}.
Further details regarding the "location of the object," "speed of our car," "distance to the object," and "our car's direction" are fed into (part 2 in Figure \ref{fig:workflow}) the large language model for mathematical calculations within CARLA (part 3 in Figure \ref{fig:workflow}). 
After formulating these calculations based on overcoming weather conditions, precise control values for brake and speed (part 4 Figure \ref{fig:workflow}) are generated and sent back to both CARLA and our ego car. We developed nine different situations by extending prompting and giving as input to GPT-4, from common sense to hybrid reasoning inside CARLA. 
When compared to ground truth, the evaluation showed that LLM can adeptly consider different data at the same time and make decisions in complex autonomous system scenarios.

\section{Related work}
Large language models have showcased remarkable performance in natural language processing tasks such as news summarization\cite{text-sum,benchnews}, sentiment analysis\cite{ qinn,surveyllm},
Question-answering\cite{Bian2023} and lexicalizing of arbitrary knowledge graph triples\cite{Axelsson}.
While Large Language Models (LLMs) have shown adept at a variety of tasks involving natural language processing and have received substantial studies, on their ability to make decisions inside simulations to provide improved driving instructions to cars remains largely unexplored.
 \paragraph{\textbf{Generalization of Large Language Models in Autonomous Car Driving}}

Developing scenarios to articulate test scenarios by incorporating the traffic rules outlined in TARGET \cite{target}. 
Evaluation tests with Automated Driving Systems(ADS) scenarios showed that TARGET is capable of detecting traffic violations with the most advanced ADS. \\
The paper \cite{semantic} discusses semantic anomaly detection through object detection with YOLOv8, showcasing how Large Language Models (LLMs) can bolster anomaly identification within specific tasks.\\
The implementation of large language models (LLMs) like GPT-series and instructive language models such as ChatGPT \cite{chatgpt} provides greater chances to manage complex language instructions. 
A separate investigation \cite{drive-like-human} delves into the viability of harnessing a large language model (LLM) to comprehend the driving environment with human-like understanding. The authors assess the LLM's capabilities in logical reasoning, interpretation, and memorization of information within complex contexts~\cite{drive-like-human}.

\paragraph{\textbf{Reinforcement Learning for Autonomous Car Driving}}
The paper~\cite{hung} proposes a framework that combines human knowledge with deep reinforcement learning to train autonomous driving agents. The framework was evaluated in two simulated urban driving scenarios. The agent trained using the proposed method achieved the highest success rate and exhibited diverse and human-like driving behaviors. \\
 A real-time human-guidance-based(Hug) Deep Reinforcement learning (DRL) method was developed in the study \cite{wu2023} to train policies in an end-to-end autonomous driving scenario. An improved actor-critic architecture with revised policy and value networks was created. During the training process, humans can intervene and correct the DRL agent's irrational actions in real-time. The developed method was validated by human-in-the-loop experiments with 40 participants and compared to other advanced learning approaches.\\
Nevertheless, the exploration of hybrid reasoning for obtaining precise values for vehicle control and decision-making is not well-studied.

\section{Background}
\paragraph{CARLA} stands as an open-source simulator that democratizes the research domain of autonomous driving, constructed upon a scalable client-server architecture.
The CARLA simulation environment is a dependable workbench for testing innovative technologies that have the potential to become technology enablers for the autonomous car sector \cite{Preci}. Industrial applications often involve intricate production lines, posing challenges for conducting tests and deployments. 
The intricacy becomes especially apparent when evaluating machine-generated controls in real-world scenarios, potentially introducing risks. Simulations provide a method to test new approaches, and for our reasoning and decision-making, we employed the CARLA Simulator in this study.\\
 
\paragraph{Object detection}
One key part of intelligent transport systems (ITS) is traffic object detection. 
Within Intelligent Transportation Systems (ITS), traffic object detection is commonly classified into four distinct categories: vehicle detection, pedestrian detection, traffic sign detection, and detection of other obstacles \cite{li2023}. 
CNN-based object detection approaches are mainly divided into two kinds: single-stage and two-stage object detection methods \cite{Carranza}. 
Two-stage object detection algorithms are not fast enough for real-time applications~\cite{cao2023}. 
YOLOv8, a member of the YOLO (You Only Look Once) family of object detection models, stands out as the top-performing model in the YOLO series and represents the cutting-edge among YOLO models. Yolov8 uses a similar backbone to YOLOv5, replacing the CSP(Cross-Stage Partial) Layer with the new C2f(cross-stage partial bottleneck with two convolutions) module. The backbone consists of a CSPDarknet53 feature extractor, rather than the conventional YOLO neck architecture\cite{compareyolo}.
In this paper, we use YOLOv8 as our object detection algorithm. We utlized \footnote{https://universe.roboflow.com/alec-hantson-student-howest-be/carla-izloa} dataset for training yolo. 
\paragraph{Large language model}
It has been demonstrated that scaling up language models may predictably enhance performance and sample efficiency on a variety of downstream activities \cite{Emergent}. 
GPT-4 is a large multi-model that can analyze text and picture inputs and produce text~\cite{gpt4}. In this paper, we utilize GPT-4, a large language model, for conducting inference in our scenarios.

 \section{LLM-based reasoning}
A key objective in machine reasoning research is to develop interpretable AI systems capable of addressing problems or drawing inferences using both pre-existing knowledge and experiences, as well as information provided to them, such as facts and observations~\cite {whatreason}. 
Within this section, we will delineate our nine scenarios and clarify the three distinct types of reasoning employed: common sense, arithmetic, and hybrid reasoning, which is a combination of common sense and arithmetic.
In the subsequent part of this section, we initially present several assumptions utilized in CARLA as configurations and settings for our scenarios. Following that, we present each reasoning approach, developed under three distinct weather conditions: heavy rain, sunny, and partly sunny.
In our evaluation, we compare answers from LLMs to human-generated ground truth. 
For each reasoning approach, we provide a discussion of the strengths and weaknesses of LLM answers briefly.
 
 \subsection{Assumption} In this section, we provide a comprehensive overview of the scenarios we have meticulously designed.
 Our objective is to investigate the responses generated by the Large Language Model (LLM) under various reasoning contexts.
 Each scenario is equipped with YOLOv8 detection capabilities within the CARLA simulation environment, and all scenarios originate from identical starting points within "Town7". 
 Importantly, these scenarios lack predefined endpoints, and the vehicles' directions are governed by the CARLA simulator's dynamic settings. Furthermore, all other vehicles in the CARLA simulation operate in autopilot mode.
We implemented nine different scenarios in CARLA, and for each scenario, we obtained five responses from the Large Language Model (LLM). Among these responses, one answer for each reasoning is randomly selected for reporting. 
All the responses can be found on our GitHub repository\footnote{https://github.com/Mehdiazarafza/Hybrid-reasoning}.

To enhance our understanding of GPT-4's responses, we made specific adjustments within CARLA.
We modified the TrafficManager settings to increase the vehicle speeds and decrease the safe distance parameter to zero, ensuring a more dynamic and challenging environment.
Additionally, we introduce ten randomly spawned vehicles into the town to create a realistic and dynamic simulation for comprehensive measurement and analysis.
For each of the nine scenarios, we supply GPT-4 with specific parameters, including the current speed of our car, the detected object identified by YOLO, the confidence score associated with the detected object's class, the distance between our car and the detected vehicle, and the direction in which our car is moving. We consider vehicles that approach from the front. However, we provide the specific location of the other vehicle, indicating whether it is on the left, right, or directly in front.
In many parts of Town 7, the speed limit is 40 km/h. However, to evaluate the decisions made by the LLM, we manually increase our car's speed beyond 40 km/h in different places. 
Therefore, if the speed is below 40 km/h and a vehicle is detected, we do not consider that scenario in our analysis.
For each type of reasoning, namely arithmetic, commonsense, and hybrid reasoning, we designed three distinct scenarios including heavy rain, sunny weather, and partly sunny without cloud conditions. 
It is notable that due to an extensive number of parameters of the LLM, the response time requires a considerable amount of time. As a result, we return the first answer to CARLA control.


\subsection{Common-sense reasoning}
The ease with which human drivers make decisions is significantly influenced by commonsense knowledge and common sense reasoning. 
 Reasoning about actions is vital for humans as it aids in predicting whether a series of actions will lead to the desired goal~\cite{reaosningShailaja}, 
but this task might be challenging for autonomous automobiles.
They must simultaneously consider elements such as speed, identified objects, distances, weather, and other factors in their immediate vicinity.
For example, humans can quickly decide to slow down or stay at a higher distance under severe wet circumstances, while autonomous automobiles find it difficult to make such judgments while taking all these factors into account.
To investigate how well common sense reasoning performs in different weather circumstances, we created three different scenarios in the CARLA simulation system. 
The purpose of these scenarios was to evaluate the autonomous system's proficiency in handling the complexity of semi-real-world circumstances, especially with bad weather. 
Our objective in subjecting the system to these scenarios was to evaluate its performance and enhance its capacity to make decisions in diverse and challenging driving situations. 
\begin{figure}[ht]
  \subcaptionbox*{Sunny weather}[.45\linewidth]{%
    \includegraphics[width=\linewidth]{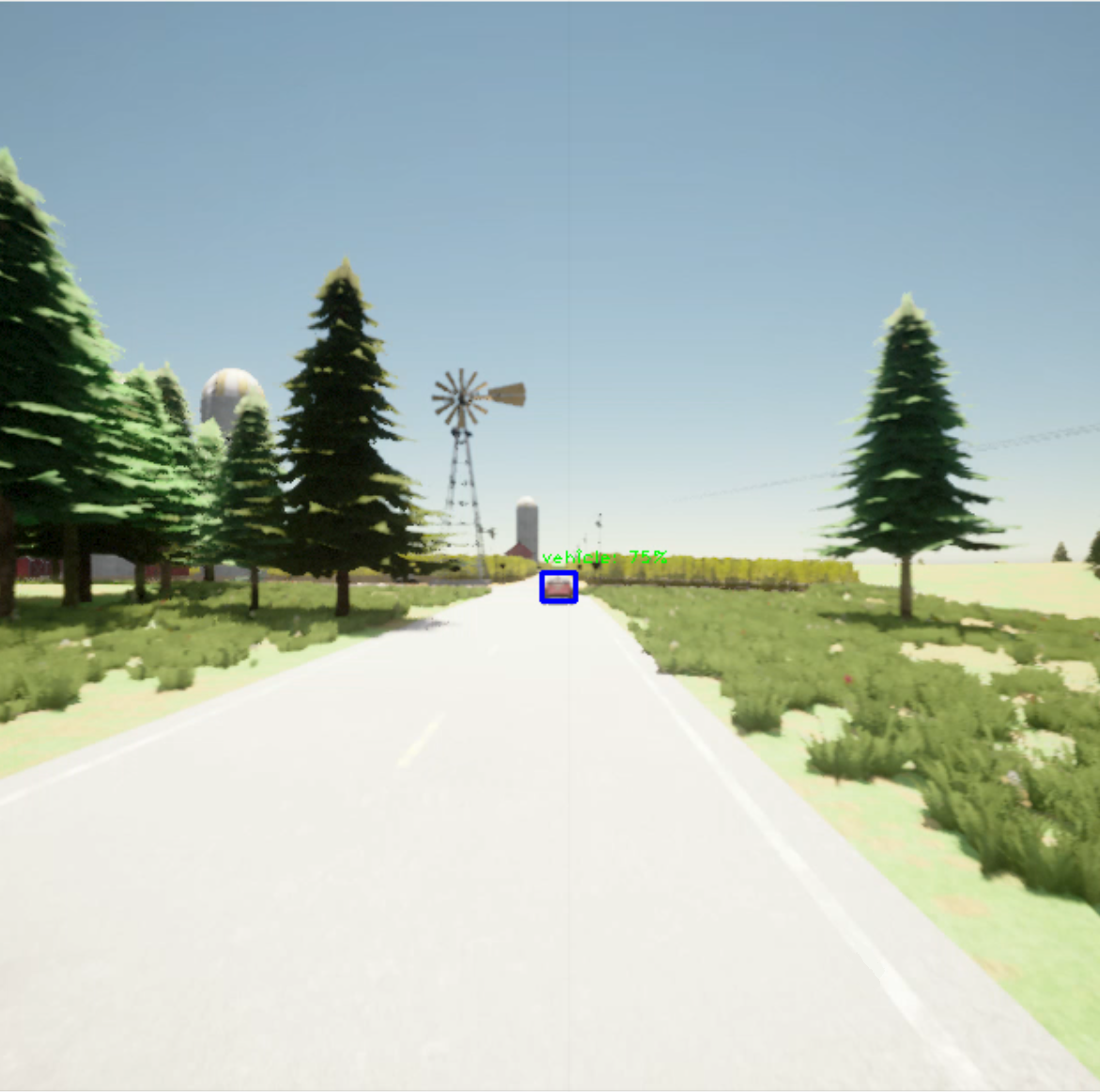}%
  }%
  \hfill
  \subcaptionbox*{Partly sunny weather}[.45\linewidth]{%
    \includegraphics[width=\linewidth]{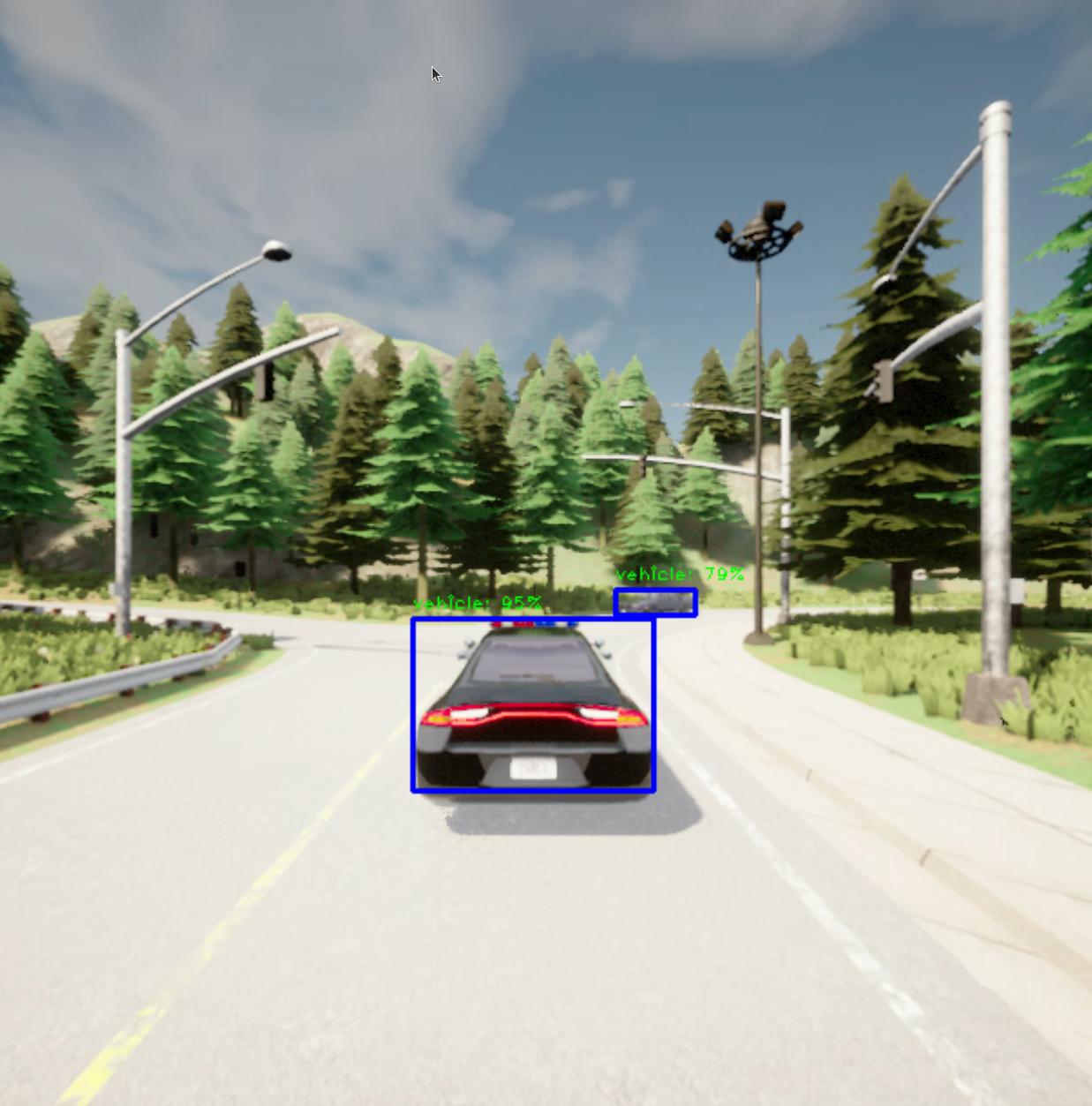}%
  }
  \caption{Detected objects in two different kinds of weather in common-sense reasoning}
  \label{fig:common-weather}
\end{figure}
For rainy conditions, the input provided to LLM for common sense reasoning is outlined in the box below. Input data from CARLA is represented within the green box. Prompt 1 has been employed as a template for other prompts, with additions and modifications. Each green box originates from CARLA, containing input sensor data and detected objects. The entire template is then used as a prompt for LLM.
We have three distinct scenarios for each weather condition, however, to report the output for each reasoning, we are presenting one of them for each reasoning, additional outputs and findings are available on our GitHub repository. Figure \ref{fig:common-weather} illustrates the detection of vehicles in 'Sunny' weather (left side) and 'Partly Sunny' weather (right side) using a randomly chosen sample input provided as text to GPT-4 in common-sense reasoning. 
\begin{tcolorbox}[
  colback=white!10,   
  colframe=blue!50,  
  arc=5mm,           
  title=Prompt 1: part of the input for rainy conditions into LLM for common sense reasoning, 
]
considering these factors and the weather conditions? 
What steps should I follow to ensure safe driving, adhere to the speed limit, and maintain a safe distance from the vehicle in front?
 I am driving a car in fully cloudy and foggy conditions without sunlight, experiencing heavy precipitation,
        at a speed of \colorbox{turqoise}{Current speed of car} kilometers per hour. My car's sensors have detected a 40 km/h speed limit sign on the right side. \\
         Additionally, the cameras have identified a \colorbox{turqoise}{Detected Object} object on the front side with a detection confidence score of \colorbox{turqoise}{class confidence of detected object},
         The object is approximately \colorbox{turqoise}{Distance} meters away, and is on the \colorbox{turqoise}{Location of vehicle} side of my car. Currently, my car is moving in the direction of \colorbox{turqoise}{My car direction}.   What actions should I take, 
                    considering these factors and the weather conditions? 
                    What steps should I follow to ensure safe driving, adhere to the speed limit, and maintain a safe distance from the vehicle?
\end{tcolorbox}
In this section, we analyze the output from the LLM in terms of common sense reasoning for one of three scenarios about "rainy" conditions. For a specific situation, in this instance, our object detection has identified a 'vehicle' with a class confidence of 91\%. The current speed of our car is 43.21 km/h, and our distance from the other vehicle is approximately 15.2 meters. We provided prompt 1 as input to the LLM and obtained the following answer. Green boxes represent input from CARLA into GPT. We provide these information, along with detected images, as sensor data to the LLM model for analysis of answers in rainy weather, this answer is randomly selected from five answers for this specific scenario.

\begin{tcolorbox}[
  colback=white!10,   
  colframe=Orange!20,  
  arc=5mm,           
  title=Part of answer from LLM for prompt 1, 
]

1. \textbf{Reduce Speed}: Given the heavy precipitation and foggy conditions, it's crucial to reduce your speed. Even though you're currently driving at 43.216779923988035 kilometers per hour, the speed limit sign detected by your car's sensors indicates a limit of 40 km/h. It's important to adhere to this limit, especially considering the poor visibility and slippery road conditions.

2. \textbf{Maintain Safe Distance}: The vehicle in front of you is approximately 15.282174193286437 meters away. Given the weather conditions, it's advisable to increase this distance to ensure you have enough time to react if the vehicle in front of you suddenly stops or slows down. 

3. \textbf{Use Fog Lights}: If your car is equipped with fog lights, turn them on. They can help improve visibility in foggy conditions.

4. \textbf{Stay Alert}: Keep an eye on the vehicle in front of you and the road conditions. Your car's sensors have a high confidence score of 91 in detecting the vehicle in front of you, but it's always good to double-check.

5.\textbf{ Use Windshield Wipers and Defrosters}: To improve visibility, make sure your windshield wipers are working effectively to clear the precipitation. Also, use your car's defrosters to prevent the windows from fogging up.

6.\textbf{ Stay in Your Lane}: As your car is moving in the right direction, ensure you stay in your lane unless you need to change lanes for any reason. If you do need to change lanes, do so carefully and only when it's safe.

7. \textbf{Be Prepared to Stop}: Given the weather conditions, be prepared to stop if necessary. This means keeping your foot near the brake pedal so you can react quickly if needed.

8. \textbf{Avoid Sudden Movements}: Try to avoid sudden steering, braking, or accelerating, as these can cause your car to skid in slippery conditions.

\end{tcolorbox}
The driving instructions provided are generally accurate and informative. They cover the most important safety considerations for driving in heavy rain, such as reducing speed, increasing following distance, using windshield wipers and defrosters, and driving smoothly. However, there are areas where the instructions could be improved.
First, the instructions could be more specific about how much to reduce speed and increase the following distance, we aim to address this through arithmetic reasoning in the next section. 
Secondly, instead of simply suggesting "increasing following distance," provide a specific distance, such as "increase following distance to four to five car lengths."
Thirdly, the provided instructions are so general that the advice seems impersonal and lacking in specific guidance.
\section{Arithmetic reasoning}
In autonomous car driving, understanding mathematics and applying physical laws are crucial. 
The main and continuous 
technique for understanding mathematics and resolving mathematical issues is reasoning problems \cite{Napitupulu2017, Lestari}.
The previous LLM response referred to the necessity of reducing speed, yet it did not offer a concrete recommendation for the magnitude of speed reduction.
 This lack of specificity is a limitation of the previous answer. 
 In this subsection, we address this limitation by providing a more precise recommendation for reducing speed with mathematical formulation. In arithmetic reasoning, we extend the prompt 1 input with more precise questions to elicit more mathematical formulations. In this instance of LLM's output for arithmetic reasoning, we are considering the results of 'heavy rainy' scenarios for arithmetic reasoning. 
 \begin{tcolorbox}[
  colback=white!10,   
  colframe=blue!50,  
  arc=5mm,           
  title=Prompt 2: input for LLM in arithmetic reasoning with heavy rainy weather, 
]
        What actions should I take, 
considering these factors and the weather conditions? 
What steps should I follow to ensure safe driving, adhere to the speed limit, and maintain a safe distance from the vehicle in front?
 I am driving a car in fully cloudy and foggy conditions without sunlight, experiencing heavy precipitation,
        at a speed of \colorbox{turqoise}{Current speed of car} kilometers per hour. My car's sensors have detected a 40 km/h speed limit sign on the right side. \\
         Additionally, the cameras have identified a \colorbox{turqoise}{Detected Object} object on the front side with a detection confidence score of \colorbox{turqoise}{class confidence of detected object},
         The object is approximately \colorbox{turqoise}{Distance} meters away, and is on the \colorbox{turqoise}{Location of vehicle} side of my car. Currently, my car is moving in the direction of \colorbox{turqoise}{My car direction}.  
         Could you provide the exact speed I should maintain for every subsequent second?
        provide precise mathematical and physical formulations that you will
          use to calculate this, step by step.
           Additionally, based on these calculations,
           specify precisely how much I should decrease or increase the speed of the car.
           
\end{tcolorbox}
\begin{tcolorbox}[
  colback=white!10,   
  colframe=Orange!20,  
  arc=5mm,           
  title=Part of answer from LLM for prompt 2, 
]

Given the conditions and the information provided, here are the actions you should take:

1. \textbf{Slow down}: Given the foggy conditions and the presence of a bike in front of you, it's important to reduce your speed. This will give you more time to react if the bike suddenly changes direction or stops. 

2. \textbf{Adhere to the speed limit}: Your car's sensors have detected a 40 km/h speed limit sign. You should reduce your speed to comply with this limit.

3. \textbf{Keep a safe distance from the bike}: The bike is approximately 18.2 meters away from your car. You should maintain a safe distance from it to avoid a collision.

Now, let's calculate the exact speed you should maintain for every subsequent second. First, let's convert the speed limit from km/h to m/s. 

\begin{equation}
    1\text{ km}/\text{h} = 0.277778\text{ m}/\text{s}
\end{equation}

So,\begin{equation}
    40 \text{km}/\text{h} = 40 \times 0.277778  =  11.1111\text{ m}/\text{s}
\end{equation} 

You are currently driving at 45.25227775733768 \text{km}/\text{h}, \\
which is \begin{equation}
    45.25227775733768 \times 0.277778 = 12.5694\text{ m}/\text{s}
\end{equation}

To reduce your speed from $12.5694 \text{ m}/\text{s}$ to $11.1111 \text{ m}/\text{s}$, you need to decelerate. \\The rate of deceleration depends on various factors such as the condition of the road, the type of vehicle, etc. \\
Let's assume a moderate deceleration rate of $1 \text{ m}/\text{s}^2$.\\
The formula to calculate the final speed is:
\begin{equation}
v = u - a \cdot t
\end{equation}
where:
v is the final speed,
u is the initial speed,
a is the acceleration (in this case, deceleration, so it's negative), and
t is the time.
So, for every subsequent second, your speed should be:\\

\begin{equation}
v= 12.5694 - 1 \cdot t
\end{equation}
For t = 1, 2, 3, ..., until v reaches 11.1111 m/s.\\
Based on these calculations, you should decrease your speed by approximately 1.4583 m/s to comply with the speed limit. This deceleration should be done gradually over a period of time to ensure a smooth and safe drive.
\end{tcolorbox}

The answer is mathematically correct with 1.4584 m/s   

Although the provided answer is mathematically correct, it lacks practical application.

It provides a formula for calculating the speed but does not offer clear guidance on how to apply it in run-time and simulated situations. 
The answer also does not specify the exact amount of braking or speed control required.
To address these challenges, we will explore the concept of hybrid reasoning in the following section.
\subsection{Combination of common-sense and arithmetic reasoning}
\begin{figure}[htbp]
\includegraphics[width=1\linewidth, height=10cm]{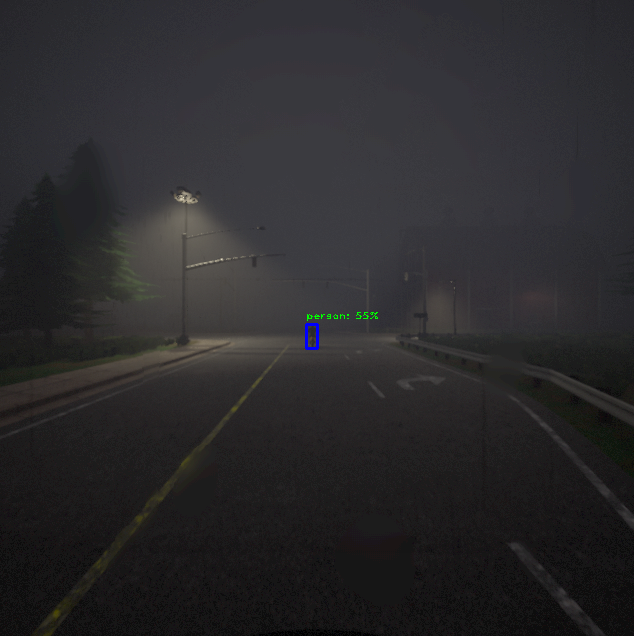} 
\centering
\caption{Detected 'Person' in front of the car}
\label{fig:person}
\end{figure}

This part focuses on hybrid reasoning, which is a complicated combination of mathematical and commonsense reasoning where the computing power of LLM is used for the complexity of run-time circumstances. We continue to feed LLM the same set of data.
But our goal is not only to understand the situation and weather conditions; 
we are now entering the domain of making decisions and returning this decision as precise into CARLA car control. 
Beyond comprehending the circumstances, we also want to compute exact brake and speed values. 
This dynamic involves various factors, including our vehicle's speed, the objects identified by our sensors, their positions relative to our car, our movement's direction, the distances involved, and the established speed limitations.
The union of these factors forms an intricate pattern that requires accurate calculations, decision-making, and a comprehensive understanding.
This scenario involves a pedestrian standing in the center of the road, directly ahead of our vehicle. The weather is severely foggy and rainy. Figure \ref{fig:person} illustrates the identified pedestrian.

Prompt 3 is an extended version of Prompt 2, augmented with additional information such as speed, detected objects from YOLOv8, and other information from Prompt 1. Following prompt 3 provided to GPT-4, we receive the corresponding output for hybrid reasoning, as presented in the box below. In this reasoning, our objective is to compile a comprehensive list of control parameters, focusing on brake and speed control. It's important to note that this represents a basic control implementation for an initial second, as our primary focus is showcasing hybrid reasoning rather than fully controlling the entire system. 
\begin{tcolorbox}[
  colback=white!10,   
  colframe=blue!50,  
  arc=5mm,           
  title=Prompt 3: Part of input for rainy conditions into LLM for hybrid reasoning, 
]

 What actions should I take?
         Could you provide the exact speed I should maintain for every subsequent second?
        provide precise mathematical and physical formulations that you will
          use to calculate this, step by step.
           Additionally, based on these calculations,
           specify precisely how much I should decrease or increase the speed of the car.
        I have control over both the throttle and the brake for speed control, each ranging from 0 to 1. To facilitate this, I request two lists:

        1. SPEED\textunderscore LIST: Specify the exact amount of throttle that should be applied at each step.
        2. BRAKE\textunderscore LIST: Specify the exact amount of brake that should be applied at each step.
        Provide the speed and brake information in the form of lists, named SPEED\textunderscore CONTROL and BRAKE\textunderscore CONTROL, respectively.
\end{tcolorbox}


\begin{tcolorbox}[
  colback=white!10,   
  colframe=Orange!20,  
  arc=5mm,           
  title=Part of answer from LLM for rainy scenario, 
]

The first action you should take is to slow down to the speed limit of 40 km/h. You should also be prepared to stop or slow down further if the person in front of you is in your path. 

To calculate the exact speed you should maintain for every subsequent second, we need to know the current speed, the desired speed, and the time it takes for your car to decelerate. 

Let's assume that your car decelerates at a constant rate. The formula for deceleration is:
 \begin{equation} a = \frac{v_f - v_i}{t}\end{equation}
where:
    a is deceleration,
    v\textunderscore f is the final velocity (desired speed),
    v\textunderscore i is the initial velocity (current speed), and
    t is time. Plugging in the given values:
    \begin{equation}
a= \frac{(40 \text{ km}/\text{h})-(45.22770823152422\text{ km}/\text{h})}{t}
\end{equation}
We need to convert km/h to m/s to make the units consistent. 1 km/h is approximately 0.277778 m/s. So:
\begin{equation}
a = \frac{(40\cdot0.277778\text{ m}/\text{s}) - (45.22770823152422\cdot0.277778\text{ m}/\text{s})}{t} = \frac{(11.1111\text{ m}/\text{s})-(12.5556\text{ m}/\text{s})}{t}\end{equation}\begin{equation}
    \frac{-1.4444 \text{ m}/\text{s}^2}{t}
\end{equation}
Assuming it takes 5 seconds to decelerate to the speed limit, we can solve for a:  \begin{equation} a= \frac{-1.4444\text{ m}/\text{s}^2}{5\text{ s}} = -0.28889999999999993\text{ m}/\text{s}^2
\end{equation} This means you need to decelerate at a rate of\begin{equation}
a = 0.2889\text{ m}/\text{s}^2
\end{equation} to reach the speed limit in 5 seconds. To calculate the exact speed for every subsequent second, we can use the formula:
\begin{equation}
v = v_i + a \cdot t 
\end{equation} For each second, we can calculate the speed as follows: \begin{equation}
v_1 = 45.22770823152422 \text{ km}/\text{h} - 0.2889 \text{ m}/\text{s}^2 \cdot 1\text{ s} =  44.9388 \text{ km}/\text{h} \end{equation}
\begin{equation}v_2 = 44.9388 \text{ km}/\text{h} - 0.2889 \text{ m}/\text{s}^2 \cdot 1 \text{ s} =44.6499 \text{ km}/\text{h}\end{equation}
\begin{equation}v_3 = 44.6499 \text{ km}/\text{h} - 0.2889\text{ m}/\text{s}^2 \cdot 1\text{ s} = 44.3610\text{ km}/\text{h} \end{equation}
The SPEED\textunderscore LIST would be [44.9388, 44.6499, 44.3610, 44.0721, ...].\\\\
The BRAKE\textunderscore LIST would depend on the specifics of your car, such as its weight and the effectiveness of its brakes.
However, assuming that a brake value of 1 results in a deceleration of $a = 1\text{ m}/\text{s}^2$,
you would need to apply a brake value of 0.2889 for each of the 5 seconds.\\\\ So the BRAKE\textunderscore LIST would be [0.2889, 0.2889, 0.2889, 0.2889, 0.2889]
\end{tcolorbox}

This answer is comprehensive and provides the necessary information for a driver to safely slow down to the speed limit for the first 5 seconds. It also uses consistent units and takes into account the factors that might affect the deceleration rate, such as the weight of the car and the effectiveness of the brakes. This answer considers:  
\begin{itemize}
   
    \item \textbf{Uniform deceleration}: This is the assumption that the car decelerates at a constant rate. This assumption is not always accurate, but it is a reasonable approximation for many situations.
    \item \textbf{Relationship between speed and time}: The speed of an object is approximately equal to its initial speed plus the acceleration multiplied by the time.

\end{itemize}
\section{Evaluation}



\begin{figure}[htbp]
\includegraphics[width=1\linewidth]{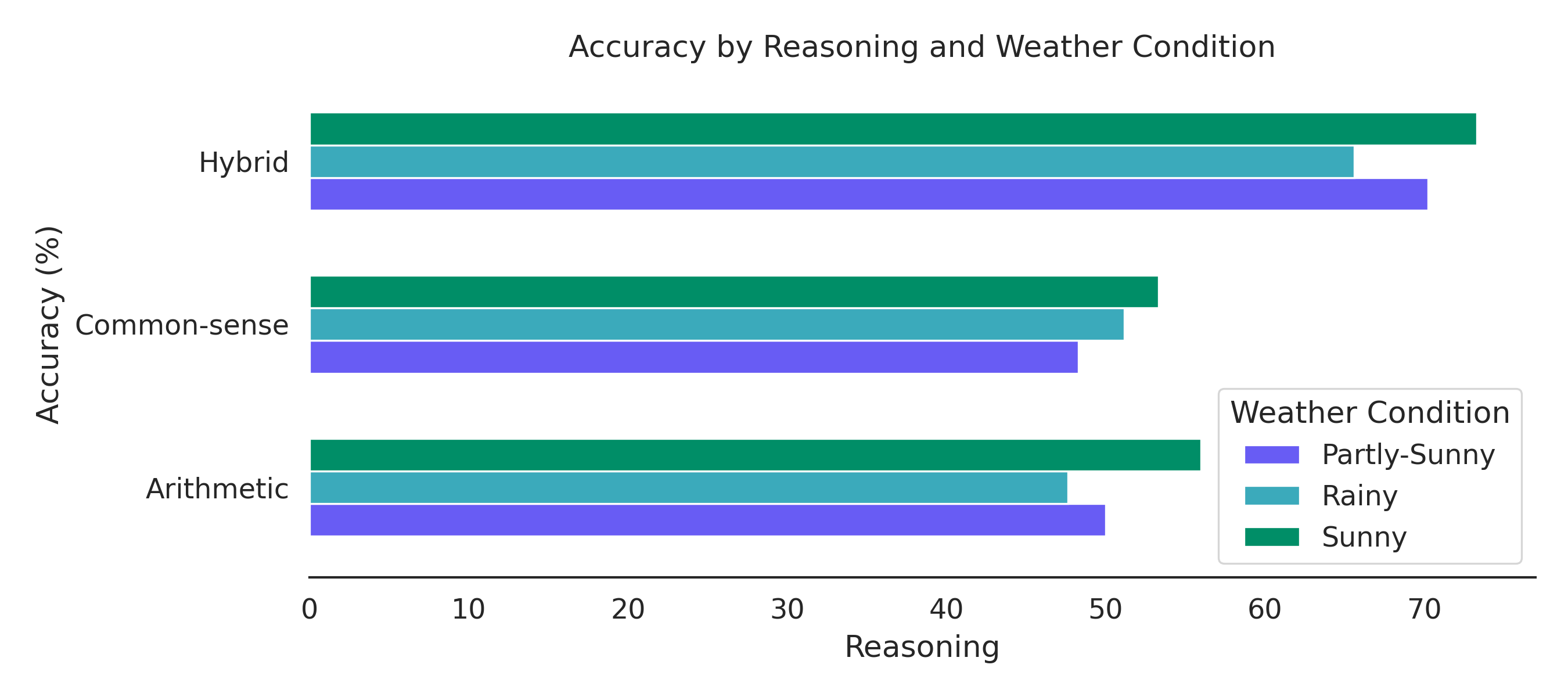} 
\caption{Comparing accuracy in different weather conditions}
\label{fig:compare}
\end{figure}

This section involves a generic comparison of events, with the main goal being to evaluate if decisions from LLM were made correctly. 
We are taking into consideration factors such as the accuracy of the formulation, and ensuring that the answer is specific rather than overly general. This detailed list of total answers, correct, incorrect answers, and their corresponding reasons is available available on GitHub \footnote{https://github.com/Mehdiazarafza/Hybrid-reasoning/tree/main/Evaulation}.
for reference. In the GitHub evaluation folder, within the 'common-sense' subdirectory, each sentence and instruction is counted as a response. For 'Arithmetic' and 'Hybrid', every formulation and correctness of all steps are considered in the count. To provide clarity on the total number of answers, a label named 'Counted Responses' is used for these two categories.\\
Subsequently, we used accuracy, and it is calculated as:
\begin{equation}
\text{Accuracy} = \frac{\text{Number of Correct Predictions}}{\text{Total Number of Predictions}}
\end{equation}
The results are illustrated in the figure \ref{fig:compare}, serving as a measurement. 

\begin{figure*}[!t]
\centering
\subfloat[]{\includegraphics[width=2.1in]{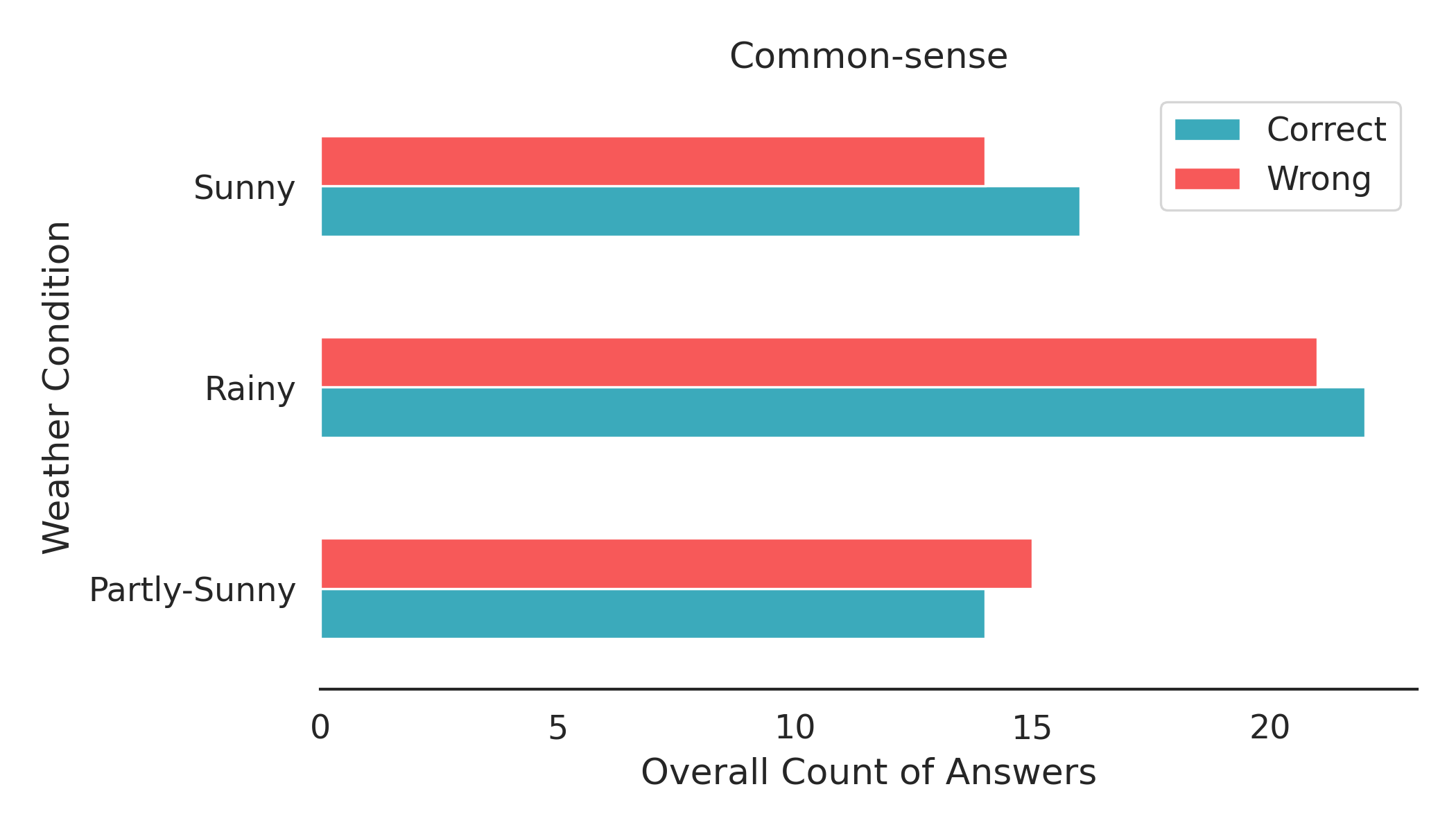}%
\label{common_result}}
\hfil
\subfloat[]{\includegraphics[width=2.1in]{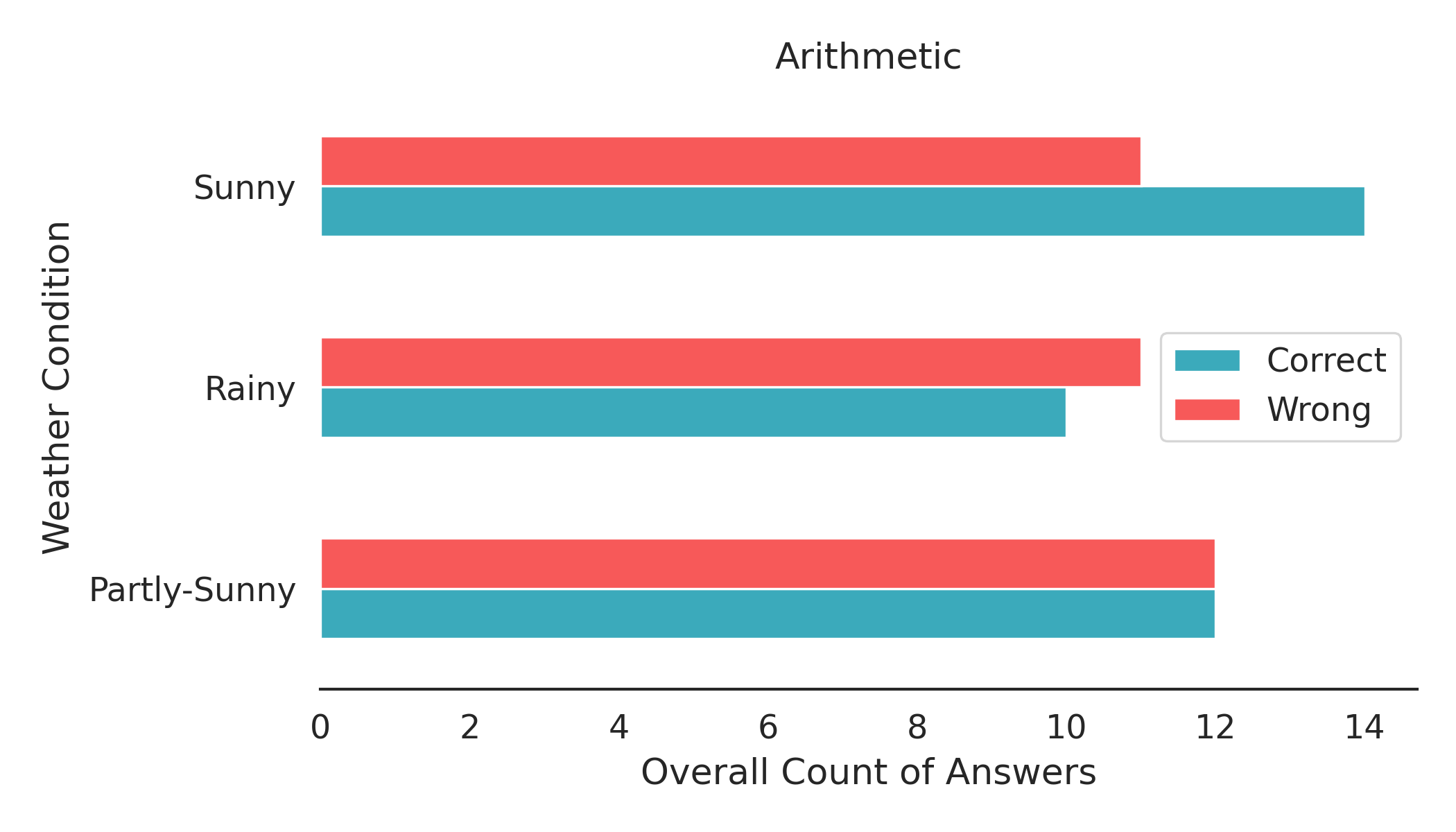}%
\label{arth_result}}
\hfil
\subfloat[]{\includegraphics[width=2.1in]{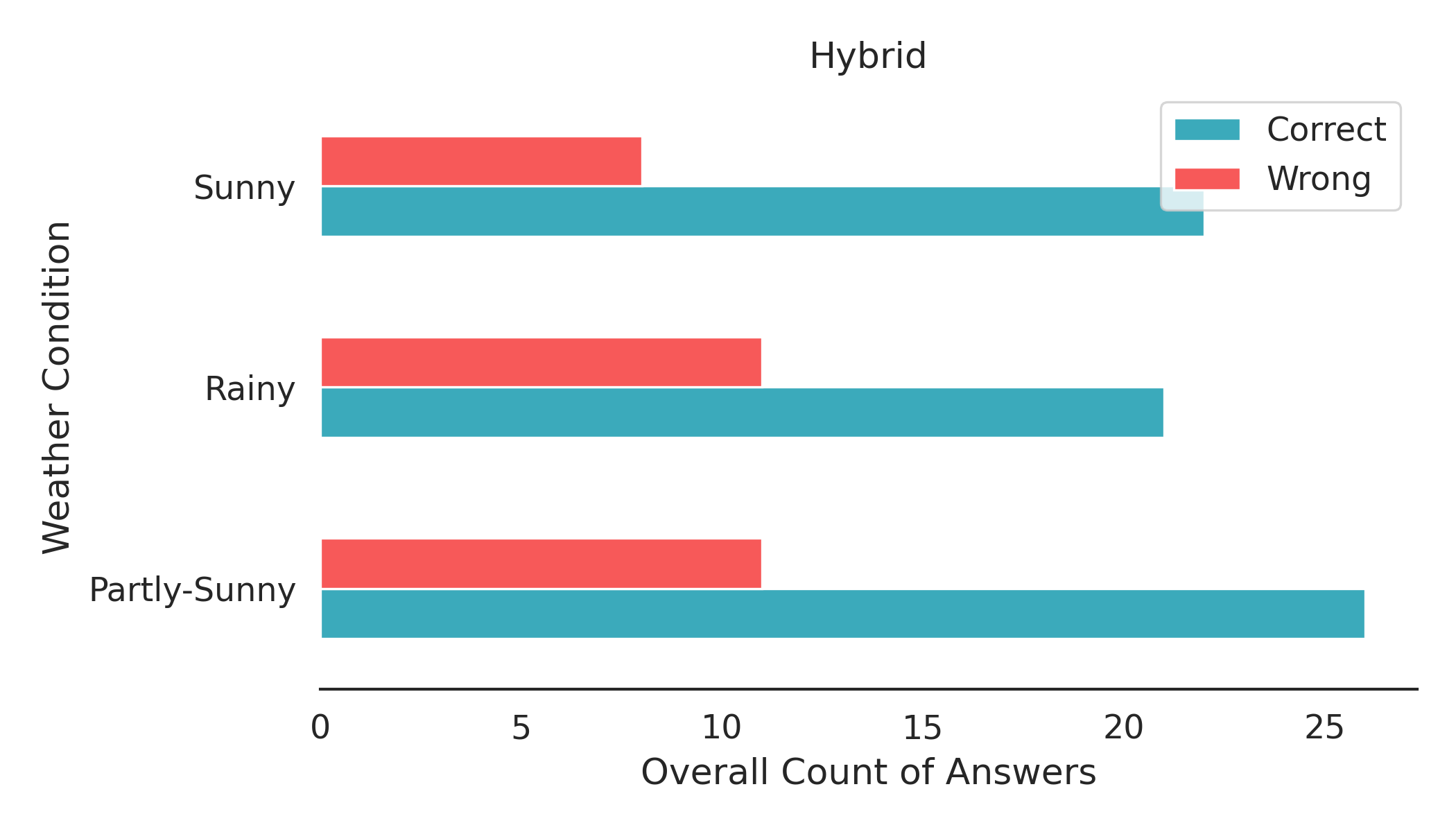}%
\label{hybrid_result}}
\caption{Comparison of Number of Correct and Incorrect Answers Using Three Methods a) Common-Sense b) Arithmetic    c) Hybrid}
\label{fig:compare_result}
\end{figure*}

As can be seen from Figure \ref{fig:compare}, the common-sense  solutions from LLM are extremely general and give general answers to particular automotive circumstances. The accuracy exceeds 53\% in sunny weather, similar to other reasoning methods that also perform well under such conditions. However, accuracy drops in 'partly sunny' weather, with the total number of answers reducing to 29 compared to 30 in 'sunny' weather, and the number of wrong answers increasing from 14 to 15. In the case of 'rainy' weather, the total number of answers gradually increases to 43, which is reasonable given the challenging weather conditions. Figure \ref{fig:compare_result} shows the number of correct and incorrect answers for all scenarios. Despite its general nature, common sense reasoning considers bad weather and provides more guidance in such situations. \\In the case of arithmetic reasoning, it performs better in 'sunny' and 'partly sunny' conditions, but its accuracy drops below 50 percent in rainy conditions. This decrease in accuracy is likely due to the neglect of factors such as visibility impact and reaction time in mathematical calculations. \\Hybrid reasoning demonstrates a notable increase in accuracy compared to the previous reasoning methods across all weather conditions. Moreover, the number of responses and correct calculations increases compared to arithmetic reasoning. Hybrid reasoning achieves an accuracy, on average, exceeding 65\% across all scenarios. Answers are improving as they become more specific and less general. This demonstrates that if we provide specific instructions to LLM, it can adjust its answers to provide precise values for autonomous car driving. This capability can be particularly helpful for autopilot systems in adverse weather conditions.
\section{Conclusion}
To sum up, our research investigated hybrid reasoning, combining mathematical reasoning with common sense reasoning in the dynamic environment of the CARLA vehicle simulator. 
We developed nine unique scenarios in CARLA, covering a range of weather conditions such as "sunny," "partly sunny," and "heavy rainy". 
We utilized Large Language Models (LLMs) with inputs that were obtained from a combination of image-detected objects and sensors data.
These inputs encompassed parameters such as object distance, current car speed, ego car direction, object location, and detected objects.
Our evaluation involved accuracy metrics, computed by dividing the number of LLM-generated answers across different situations by the number of correct answers as determined by human-generated ground truth.

The results
are underscored the effectiveness of hybrid reasoning with LLM, particularly in calculating precise brake and throttle
values over a 5-second interval, demonstrating notable accuracy and efficacy.
This information illustrates the capacity of LLMs to enhance the autonomous driving experience, particularly in bad
weather conditions, playing a pivotal role in decision-making during adverse weather conditions. In future research, we
may explore the possibility of reducing the number of input parameters and optimizing them in datasets to have step-by-step
calculation and response from LLM.
\bibliographystyle{unsrt}  


\end{document}